%% file: main.tex
\documentclass[runningheads]{llncs}

\usepackage{custom}

\begin{document}
\maketitle

\begin{abstract}
  Vision-Language Models (VLMs) have shown impressive performance in vision tasks, but adapting them to new domains often requires expensive fine-tuning.
  Prompt tuning techniques, including textual, visual, and multimodal prompting, offer efficient alternatives by leveraging learnable prompts.
  However, their application to Vision-Language Segmentation Models (VLSMs) and evaluation under significant domain shifts remain unexplored.
  This work presents an open-source benchmarking framework, \textit{TuneVLSeg}, to integrate various unimodal and multimodal prompt tuning techniques into VLSMs, making prompt tuning usable for downstream segmentation datasets with any number of classes.
  \textit{TuneVLSeg} includes $6$ prompt tuning strategies on various prompt depths used in $2$ VLSMs totaling of $8$ different combinations.
  We test various prompt tuning on $8$ diverse medical datasets, including $3$ radiology datasets (breast tumor, echocardiograph, chest X-ray pathologies) and $5$ non-radiology datasets (polyp, ulcer, skin cancer), and two natural domain segmentation datasets.
  Our study found that textual prompt tuning struggles under significant domain shifts, from natural-domain images to medical data.
  Furthermore, visual prompt tuning, with fewer hyperparameters than multimodal prompt tuning, often achieves performance competitive to multimodal approaches, making it a valuable first attempt.
  Our work advances the understanding and applicability of different prompt-tuning techniques for robust domain-specific segmentation.
  The source code is available at \url{https://github.com/naamiinepal/tunevlseg}.
\end{abstract}

\section{Introduction}

\begin{figure}[t]
    \centering
    \includegraphics[width=\linewidth]{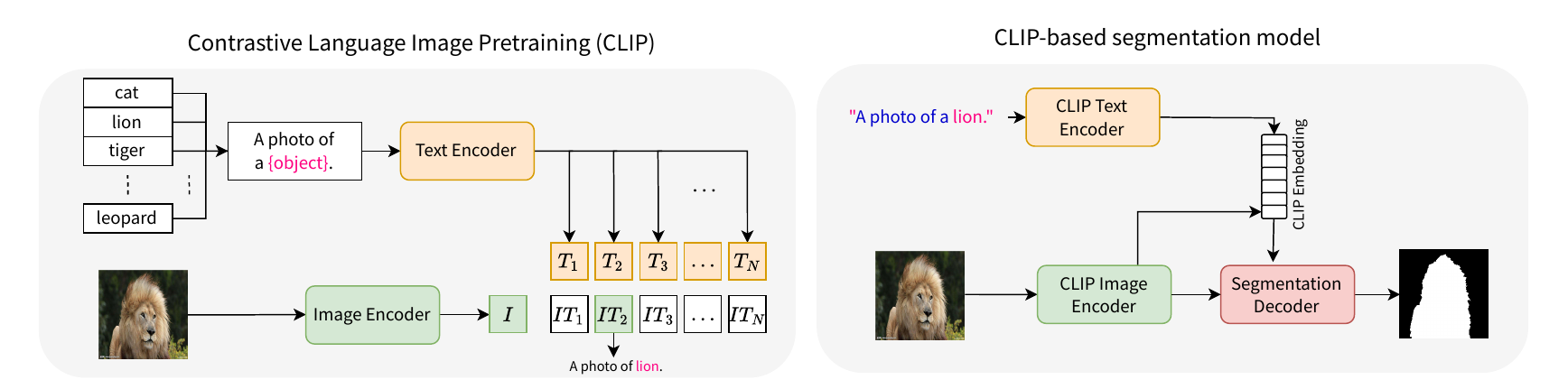}
    \caption{Manual prompting in CLIP and CLIP-based segmentation models.}
    \label{fig:prompt-tuning-overview}
\end{figure}

Segmenting the anatomical and pathological structures in medical images is crucial for computer-aided diagnosis, prognosis, and surgery planning.
Recent deep-learning-based segmentation models have shown excellent performance on curated datasets, but lack generalization across anatomies and image modalities as they are typically trained on a limited set of anatomies and modalities or fine-tuned using pretrained weights from models trained on natural images like ImageNet \cite{deng2009imagenet}.
Recent advancements in foundational vision language models (VLMs) \cite{radford2021learning,singh2022flava,zhai2022lit,jia2021scaling,zhang2022contrastive,li2021supervision}, which leverage the image and text data, have gained significant attention from the research community due to their generalization capabilities across new dataset and vision tasks.
These models have been adopted for segmentation tasks in natural images \cite{rao2022denseclip,zhou2023zegclip,wang2022cris,luddecke2022image}, and show excellent generalization for segmentation as well.
These vision-language segmentation models (VLSMs) use the pretrained encoders of the VLMs with an added segmentation decoder.
The surprising generalization ability of VLSMs stems from the language supervision provided by text inputs, known as \textit{prompts}.
During inference, prompts like ``\texttt{a photo of a \{\textit{object}\}}'' supply auxiliary information along with image embeddings to identify the target class, and high-quality prompts are crucial for enhancing VLSM's performance \cite{jin2022good,poudel2023exploring}.

Extending VLSMs to medical image segmentation tasks presents challenges, particularly in designing effective prompts for VLSMs pretrained on natural images \cite{poudel2023exploring}.
This often results in suboptimal performance for medical image segmentation, thus requiring fine-tuning on medical datasets \cite{poudel2023exploring,goyal2023finetune,zhao2023clip,shrestha2023medical,zhang2024segment}.
Given the massive scale of these models and the scarcity of large labeled medical datasets, fine-tuning VLSMs for medical datasets is often infeasible.
One of the efforts to extend these models to new domains with fewer data and computational requirements is prompt tuning.
It is robust to noisy labels \cite{wu2023prompt}, making this strategy favorable to extend VLMs pretrained on natural images to medical datasets.
The prompt tuning strategies (also known as \textit{context learners}) can be extended to VLSMs as they share a similar architecture with VLMs, which are usually composed of text and vision encoders, with an added segmentation decoder (see \cref{fig:prompt-tuning-overview}).
Prompt tuning addresses the challenge of hand-engineering prompts by introducing learnable prompts (also called \textit{context vectors}), thus adapting VLMs to new datasets by optimizing only these context vectors \cite{zhou2022learning}.
Extending prompt tuning to the segmentation task has its caveats.
As illustrated in \cref{fig:prompt_tuning_methods}, the context vectors can be introduced at text \cite{zhou2022conditional,zhou2022learning}, vision \cite{jia2022visual}, or both inputs \cite{zang2022unified,khattak2023maple}.
Additionally, these vectors can be injected at different depths at the encoders, and with an added decoder for segmentation tasks, these challenges are further escalated.
However, most studies on prompt tuning have focused primarily on image classification, limiting insights on the prospect of prompt tuning on downstream image segmentation tasks.


In this work, we propose an open-source benchmark framework, \textit{TuneVLSeg}, incorporating unimodal and multimodal prompt tuning methods for a principled way of evaluating prompt tuning for different class-agnostic VLSMs.
We study the effects of adding the context vectors at multiple depths at both image and text encoders.
We also evaluate the performance of the $6$ prompt tuning methods on adapting $2$ pretrained VLSMs to $2$ natural and $8$ medical segmentation datasets.
Our benchmark framework can be extended to adapting other class-agnostic VLSMs using new prompt-tuning methods to other segmentation datasets with little to no effort.

Our major contributions encompass the following points:
\begin{itemize}
    \item A principled way of evaluating different unimodal and multimodal prompt-tuning strategies for segmentation tasks. 
    \item A modifiable and reusable benchmark framework, \textit{TuneVLSeg}, leveraging pretrained VLSMs to fine-tune them to target segmentation tasks with prompt-tuning.
    \item Comparing the effectiveness of different unimodal and multimodal prompt-tuning in natural and medical segmentation datasets across multiple radiology and non-radiology images.
\end{itemize}

\section{Related Work}

\subsection{Vision-Language Models}

Vision-Language Models (VLMs) incorporate representation modules to connect image and text features, enabling their use in various vision-language tasks such as visual question-answering (VQA), image segmentation, image-text retrieval, object detection, and phrase grounding.
Bordes \etal \cite{bordes2024introduction} have categorized VLM training into four distinct classes.
\textit{Contrastive VLMs } \cite{radford2021learning,zhai2023sigmoid,jia2021scaling,li2021supervision} learn to project text and vision features onto the same embedding space to establish equivalence among the modalities.
\textit{VLMs with masking objectives} \cite{singh2022flava,kwon2023masked,chen2020uniter} are trained to guide models to predict missing image patches or text tokens for establishing similarity between both modalities.  
\textit{Generative models} \cite{rombach2022high,yu2022coca} are optimized for generating images using the text representation and vice-versa.
\textit{VLMs with pretrained backbone} \cite{zhai2022lit,zhu2024minigpt,tsimpoukelli2021multimodal} train a mapping network between pretrained encoders, thus, requiring fewer compute.


\subsection{Vision-Language Segmentation Models}

VLMs' capacity to capture multimodal information has led to their widespread use as a backbone for segmentation tasks.
Vision-Language Segmentation Models (VLSMs) incorporate a vision-language decoder on top of the pretrained VLM encoders to capture information from both text and vision branches.
DenseCLIP \cite{rao2022denseclip} employs a vision-language decoder atop CLIP encoders, utilizing pixel-text score maps to guide the learning of dense prediction models.
Similarly, CLIPSeg \cite{luddecke2022image} and CRIS \cite{wang2022cris} provide zero-shot segmentations by predicting pixel-level activations for the given text or image query.
ZegCLIP \cite{zhou2023zegclip} introduces tuned prompts and associates image information with text encodings before patch-text contrasting, aiming to reduce overfitting of seen classes in both inductive and transductive zero-shot settings.

\subsection{Prompt Tuning}

Prompts play a crucial role in downstream datasets \cite{yao2024cpt,jin2022good}.
However, identifying the optimal prompt for a given downstream task can be burdensome and sometimes requires prior knowledge. 
Therefore, a more effective approach is automatically learning these prompts for specific downstream tasks instead of providing manual prompts.
This paradigm of learning a task-specific prompt is known as prompt tuning, which was first used in NLP \cite{shin2020autoprompt, jiang2020can, zhong2021factual, lester2021power, li2021prefix, liu2022p, liu2023pre}. 
It was subsequently adapted for the text branch of VLMs \cite{zhou2022conditional, zhou2022learning, zhu2023prompt} and vision-only models \cite{jia2022visual, zhang2022neural, wang2022dualprompt, wang2022learning}. 
Finally, multimodal prompt tuning strategies \cite{khattak2023maple, zang2022unified} were applied to both text and vision encoders of VLMs, demonstrating superior performance compared to unimodal approaches.

Although prompt tuning strategies have been extensively studied for VLMs like CLIP \cite{radford2021learning}, few studies have focused on applying these techniques to VLSMs. 
DenseCLIP \cite{rao2022denseclip} and ZegCLIP \cite{zhou2023zegclip} introduced CoOp \cite{zhou2022learning} and VPT \cite{jia2022visual} for VLSMs, respectively.
However, the main drawback of these models is that changes in the number of classes in the segmentation task alter the number of channels in their weights.
This renders their pretrained weights unusable for downstream tasks unless the desired class(es) were present during pretraining.

\section{Revisiting CLIP}


\subsection{Text Encoding}
\label{sec:clip_text_encoding}

CLIP's text encoder tokenizes the provided text with some special tokens, generates their corresponding word embeddings, and adds the position encoding before passing them to the first transformer layer, $W_0 = [w_0^1, w_0^2, \cdots, w_0^N] \in \mathbb{R}^{N \times H_l}$.
Each $W_i$ is passed to the $(i+1)^{th}$ transformer layer $\mathcal{L}_{i+1}$, outputting $W_{i+1}$.
Mathematically,\
\begin{equation}
    W_{i+1} = \mathcal{L}_{i+1}(W_i) \qquad\qquad i = 0, 1, \cdots, K_l - 1
\end{equation}

Here, $K_l$ denotes the number of transformer layers for the text encoder.
Furthermore, the sentence embedding is extracted from the output the ultimate layer ($W_K$) as the last vector in the sequence (corresponding to the End of Sentence (EOS) token), \ie, $W_K^N \in \mathbb{R} ^ {H_l}$.
The sentence embedding in the text space is projected into the vision-language space using a learnable linear projection matrix, $\mathcal{P}_l \in \mathbb{R} ^ {H_l \times H_{vl}}$.
Mathematically,
\begin{equation}
    z_l = \mathcal{P}_l^T W_K^N \qquad\qquad z_l \in \mathbb{R}^{H_{vl}}
\end{equation}

\subsection{Image Encoding}
\label{sec:clip_image_encoding}

CLIP's image encoder splits the image into multiple patches of predefined sizes.
It projects them linearly to obtain token embeddings and adds the position encoding to them before passing to the first ViT layer, \ie, $E_0 = [e_0^1, e_0^2, \cdots, e_0^N] \in \mathbb{R}^{N \times H_v}$.
Additionally, a class (CLS) token embedding ($c_i \in \mathbb{R}^{H_v}$) is concatenated to the token embedding ($E_i$) before passing to the transformer layer $\mathcal{V}_{i+1}$,  outputting $[c_{i+1}, E_{i+1}]$.
Mathematically,
\begin{equation}
    [c_{i+1}, E_{i+1}] = \mathcal{V}_{i+1}([c_i, E_i]) \qquad\qquad i = 0, 1, \cdots, K_v - 1
\end{equation}

Here, $K_v$ denotes the number of transformer layers for the image encoder.
The class (CLS) embedding from the last layer ($c_K$) corresponds to the global embedding (or image-level embedding), which is in the image space.
The embedding is projected into the vision-language space using a learnable linear projection matrix, $\mathcal{P}_v \in \mathbb{R} ^ {H_t \times H_{vl}}$.
Mathematically,
\begin{equation}
    z_v = \mathcal{P}_v^T c_K \qquad\qquad z_v \in \mathbb{R}^{H_{vl}}
\end{equation}

\section{Revisiting Prompt Tuning}

\subsection{Textual Prompt Tuning}
To learn language-specific prompts, we assign $B$ learnable tokens to the first transformer layer for CLIP's text encoder, as $P_0 = [p_0^1, p_0^2, \cdots, p_0^B] \in \mathbb{R}^{B \times H_l}$.      
The input for the first transformer layer of the text branch is the concatenation of the learnable prompts and provided fixed text prompt, \ie, $[P_0, W_0]$ (see \cref{fig:coop_overview}).
Also, these learnable prompts can be extended to multiple layers of the text encoder, such that $J$ is the network depth up to which new prompts are learned for each layer.
This deep prompting technique helps align the intermediate layers' outputs with the first layer's output.
Mathematically,
\begin{equation}
    [\_, W_{i+1}] = \mathcal{L}_{i+1}([P_i, W_i]) \qquad\qquad i = 0, 1, \cdots, J - 1
\end{equation}

Up to $J^{th}$ transformer layer, the outputs corresponding to learnable prompts $P$ are discarded, and a new set of learnable prompts is injected for each layer.
Whereas, after the $J^{th}$ layer ($\mathcal{L}_J$), the computation is carried on similar to the one described in \autoref{sec:clip_text_encoding} as if the prompts are identical to the word embeddings.
Mathematically,
\begin{equation}
    [P_{i+1}, W_{i+1}] = \mathcal{L}_{i+1}([P_i, W_i]) \qquad\qquad i = J, J + 1, \cdots, N_l - 1
\end{equation}

When the depth of the network to which the prompt is injected is unity, \ie, $J=1$, it subsides to CoOp \cite{zhou2022learning}.
CoCoOp \cite{zhou2022conditional} extended CoOp by adding the projection of CLIP's image-level embeddings to learnable prompts.
It makes the textual prompts image-dependent compared to the instance-agnostic task-dependent prompts of CoOp.



\subsection{Visual Prompt Tuning}

Similar to textual prompt tuning, we assign $B$ learnable tokens to the first transformer layer for CLIP's vision encoder, as $\tilde{P}_0 \in \mathbb{R}^{B \times H_v}$ (see \cref{fig:vpt_overview}).
The input for the first transformer layer of ViT is the concatenation of the CLS token, fixed image token embeddings, and the learnable prompts, \ie, $[c_0, E_0, \tilde{P}_0]$.
Mathematically,
\begin{equation}
    \label{eqn:vpt_prompt_replaced}
    [c_{i+1}, E_{i+1}, \_] = \mathcal{V}_{i+1}([c_i, E_i, \tilde{P}_i]) \qquad\qquad i = 0, 1, \cdots, J - 1
\end{equation}

Similar to textual prompt tuning, up to $J^{th}$ transformer layer, the outputs corresponding to learnable prompts $\tilde{P}$ are discarded, and a new set of learnable prompts is injected for each layer.
Whereas, after the $J^{th}$ layer ($\mathcal{V}_J$), the computation is carried on similar to the one described in \autoref{sec:clip_image_encoding} as if the prompts are identical to the token embeddings.
Mathematically,
\begin{equation}
    [c_{i+1}, E_{i+1}, \tilde{P}_{i+1}] = \mathcal{V}_{i+1}([c_i, E_i, \tilde{P}_i]) \qquad\qquad i = J, J + 1, \cdots, N_v - 1
\end{equation}



\subsection{Multimodal Prompt Tuning}
\label{sec:multimodal_prompt_tuning}

\begin{figure}[t]
    \centering
    \begin{subfigure}[t]{0.33\textwidth}
        \includegraphics[width=\textwidth]{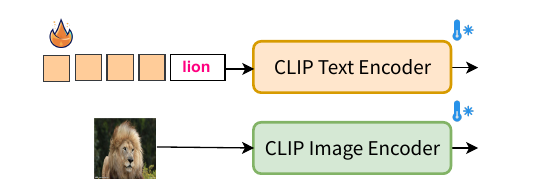}
        \caption{CoOp Overview}
        \label{fig:coop_overview}
    \end{subfigure}%
    \begin{subfigure}[t]{0.33\textwidth}
        \includegraphics[width=\textwidth]{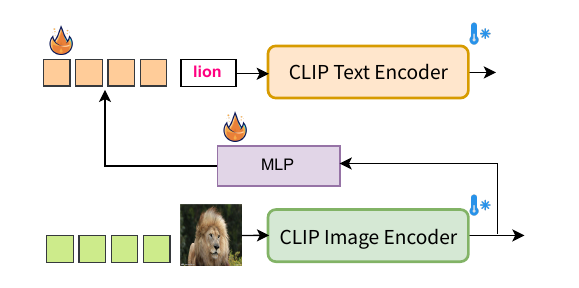}
        \caption{CoCoOp Overview}
        \label{fig:cocoop_overview}
    \end{subfigure}%
    \begin{subfigure}[t]{0.33\textwidth}
        \includegraphics[width=\textwidth]{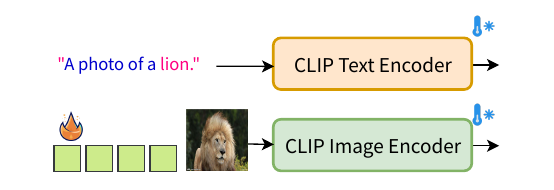}
        \caption{VPT Overview}
        \label{fig:vpt_overview}
    \end{subfigure}
    \begin{subfigure}[b]{0.33\textwidth}
        \includegraphics[width=\textwidth]{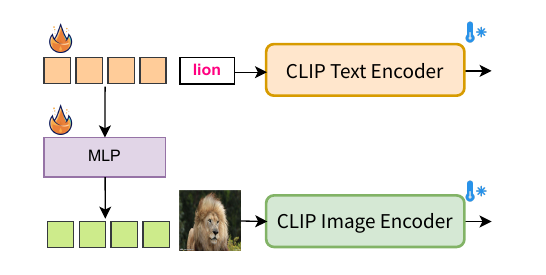}
        \caption{Maple Overview}
        \label{fig:maple_overview}
    \end{subfigure}%
    \begin{subfigure}[b]{0.33\textwidth}
        \includegraphics[width=\textwidth]{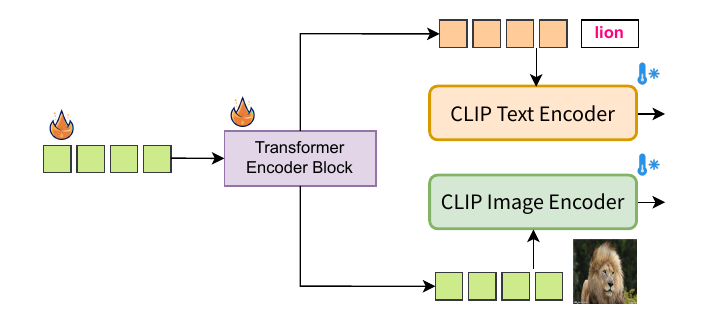}
        \caption{Shared Attention Overview}
        \label{fig:shared_attn_overview}
    \end{subfigure}%
    \begin{subfigure}[b]{0.33\textwidth}
        \includegraphics[width=\textwidth]{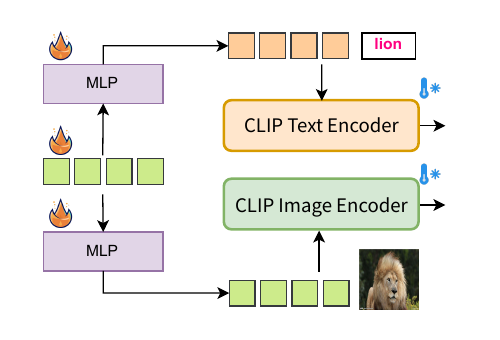}
        \caption{Shared Separate Overview}
        \label{fig:shared_separate_overview}
    \end{subfigure}%
    \caption{
    Overview of various prompt tuning methods.
    In the first row, there are unimodal prompt tuning methods and the second row shows the multimodal prompt tuning methods. 
    The prompting for only the first layer is shown here, the same concept is applicable when the prompt tuning is done for multiple transformer blocks. 
    }
    \label{fig:prompt_tuning_methods}
\end{figure}

Multimodal approaches to prompt tuning were devised to adjust the output for both the encoders.
Similar to unimodal prompt tuning, we assign $B$ tokens to each transformer layer $i$ till the prompt depth $J$ for CLIP's text and vision encoders, as $P_i \in \mathbb{R}^{B \times H_l}$ and $\tilde{P}_i \in \mathbb{R}^{B \times H_v}$, respectively.

A straightforward way to implement the multimodal prompt tuning is to learn both textual ($P_i$) and visual ($\tilde{P}_i$) prompts independently in the same training schedule.
However, studies have shown that such a naive training strategy leads to sub-optimal performance due to the lack of interaction between the vision and language branches \cite{khattak2023maple,zang2022unified}.
The interaction between the two modalities can be achieved by introducing unified prompts for each transformer layer $i$, denoted by $\hat{P}_i \in \mathbb{R}^{B \times H_u}$.
Here, $H_u$, a hyperparameter, is the dimension of the unified prompt.
We obtain visual and textual prompts from the corresponding unified prompts as follows.
\begin{equation}
    [P_i, \tilde{P}_i] = \mathcal{U}_i(\hat{P}_i)
\end{equation}

Here, $\mathcal{U}_i$ is a learnable function that transforms unified prompts $\hat{P}_i$ for each transformer layer $i$ into corresponding textual $P_i$ and visual $\tilde{P}_i$ prompts.
Some architectures of functions $\mathcal{U}_i$ are self-attention mechanisms, multi-layer perception, or a mixture of both.
For this work, we use a transformer block for each $\mathcal{U}_i$ as one of the methods of multimodal prompt tuning, named \textit{Shared Attention} (see \cref{fig:shared_attn_overview}).


One specialized case of these transformations is when the transformations are split into separate components of text and vision transformations, viz. $\mathcal{U}^l_i$ and $\mathcal{U}^v_i$ which transform the unified prompts to the corresponding text and vision spaces.
For example, the transformation function is a perceptron (an affine transformation) --- and optionally followed by an activation function.
Mathematically,
\begin{equation}
    \mathcal{U}_i(\hat{P}_i) = [\mathcal{U}^l_i(\hat{P}_i), \mathcal{U}^v_i(\hat{P}_i)]
\end{equation}
Equivalently,
\begin{equation}
    P_i = \mathcal{U}^l_i(\hat{P}_i) \qquad \tilde{P}_i = \mathcal{U}^v_i(\hat{P}_i)
\end{equation}


For these separate transformations, we use a linear layer followed by layer normalization \cite{ba2016layer} to project the unified prompts to textual and visual dimensions.
We call the method \textit{Shared Separate} (see \cref{fig:shared_separate_overview}).
A more specialized case of these separable transforms is when one of the transforms, $\mathcal{U}^l_i(\hat{P}_i)$ or $\mathcal{U}^v_i(\hat{P}_i)$, is identity.
In that case, the unified prompts are directly initialized to the space for which the transformation is identity.
Khattak \etal \cite{khattak2023maple} does the same;
they initialize the unified prompts in the text space (see \cref{fig:maple_overview}).


\section{Method}
\label{sec:method}

\begin{figure}[t]
    \centering
    \includegraphics[width=\linewidth]{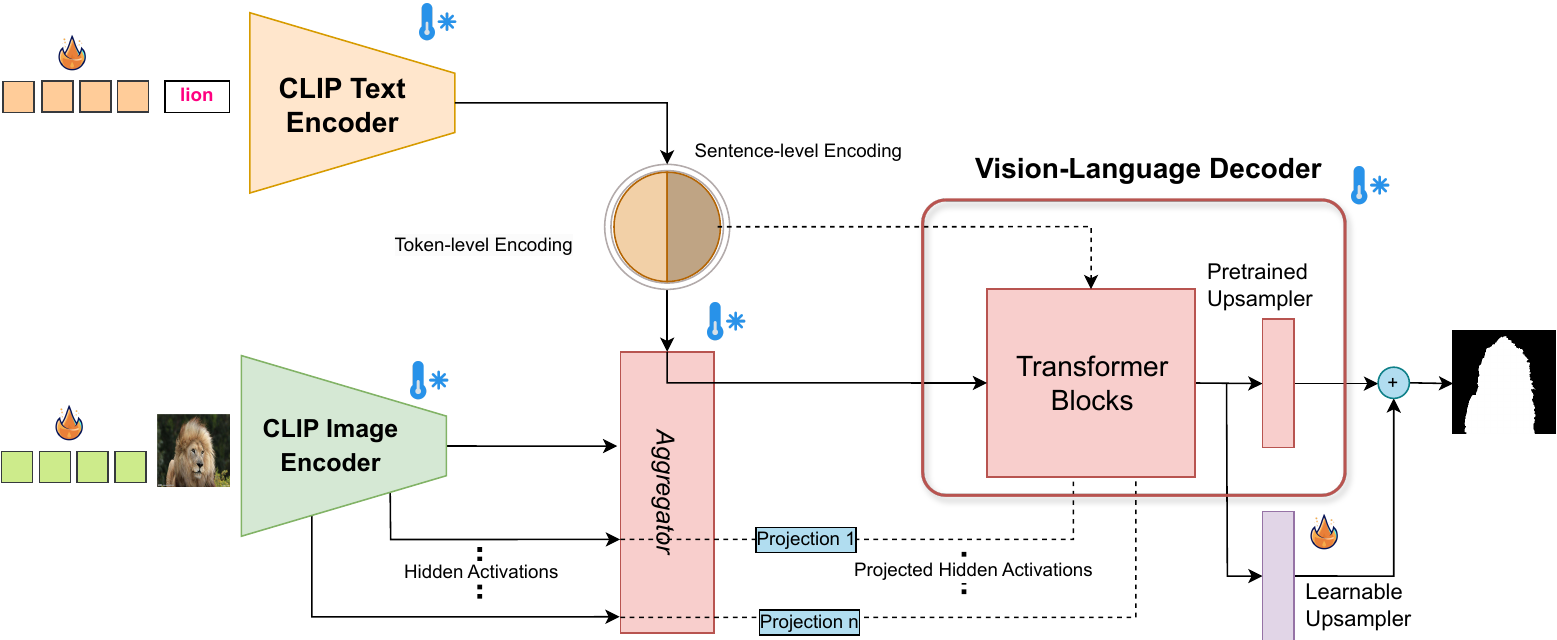}
    \caption{Multimodal Prompt Tuning Architecture.
    To simplify, the projection layers for conditioning prompts from one mode to another are not shown here.
    Likewise, for unimodal techniques, only either of the prompt modalities is fed into the model.}
    \label{fig:prompt_tuning_arch}
\end{figure}


\cref{fig:prompt_tuning_arch} shows the general architecture of our proposed framework.
For multimodal prompt tuning, the learnable prompts are injected in both the encoders whereas for unimodal prompt tuning, the prompts are injected in the encoder of the respective modality. 
Also, since all the parameters of the pretrained models are frozen, we train and store only the learnable contexts for each downstream task, thus saving computational resources and storage.


In our study, we have used two VLSMs that are pretrained on natural images viz. CLIPSeg \cite{luddecke2022image} and CRIS \cite{wang2022cris} for prompt tuning.
A key difference between CLIPSeg's and CRIS's architecture is that CLIPSeg uses ViT \cite{dosovitskiy2020image} as the vision encoder while CRIS uses ResNet \cite{he2016deep, he2019bag}.
And, while textual prompt tuning is used for both models, the visual and multimodal prompt tuning techniques are applied only to CLIPSeg \cite{luddecke2022image} because they require a ViT encoder.
Thus, we have used six prompt tuning methods for CLIPSeg --- three multimodal, one visual, and two textual, whereas two textual prompt tuning methods for CRIS.
This amounts to a total of eight combinations of prompt tuning and VLSMs for each dataset.

Furthermore, inspired by Jia \etal \cite{jia2022visual}, we have introduced a learnable upsampler (bottom right of \cref{fig:prompt_tuning_arch}) that takes the output of the penultimate layer of the decoder; its output is added to the decoder's output with learnable residual factor.
This learnable block consists of a bilinear upsampler followed by a 2D convolution layer with kernel size $5 \times 5$. 

\section{Experiments}

\subsection{Datasets}
\label{sec:datasets}

For our empirical analysis of medical datasets, we utilize eight of the datasets and their splits provided by Poudel \etal \cite{poudel2023exploring}.
These datasets include five non-radiology datasets --- Kvasir-SEG \cite{jha2020kvasir}, ClinicDB \cite{bernal2015wm}, and BKAI \cite{ngoc2021neounet,an2022blazeneo} for polyp segmentation in endoscopic images, DFU \cite{kendrick2022translating} for diabetic foot ulcer segmentation, and ISIC-16 \cite{gutman2016skin} for skin lesion segmentation --- and three radiology datasets --- BUSI \cite{al2020dataset} for breast ultrasound segmentation, CAMUS \cite{leclerc2019deep} for 2D echocardiography segmentation, and CheXlocalize \cite{saporta2022benchmarking} for chest X-ray segmentation. 
For our analysis, we only use the foreground class name to learn the prompt for each dataset automatically.

\begin{table}[t]
    \centering
    \caption{
        An overview of our datasets compared across the dimensions of category, type or modality, organ (for medical datasets), foreground classes, and their splits.
    }
    \label{tab:dataset_info}
    {\tiny
    \begin{tabular}{@{}lp{0.11\linewidth}llp{0.36\linewidth}c@{}}
        \toprule
        \textbf{Category} & \textbf{Type} & \textbf{Organ} & \textbf{Name} & \textbf{Foreground Class(es)} & \textbf{\# train/val/test} \\ \midrule
        \multirow{5}{*}{Non-Radiology} & \multirow{3}{*}{Endoscopy} & \multirow{3}{*}{Colon} & Kvasir-SEG & \multirow{3}{*}{Polyp} & 800/100/100 \\
         &  &  & ClinicDB &  & 490/61/61 \\
         &  &  & BKAI &  & 800/100/100 \\ \cmidrule{2-6} 
         & \multirow{2}{*}{Photography} & Skin & ISIC 2016  & Skin Lesion & 810/90/379 \\
         &  & Foot & DFU 2022 & Foot Ulcer & 1600/200/200 \\ \midrule
        \multirow{3}{*}{Radiology} & \multirow{2}{*}{Ultrasound} & Heart & CAMUS & Myocardium, Left ventricular, and Left atrium cavity & 4800/600/600 \\
         &  & Breast & BUSI & Benign and Malignant Tumors & 624/78/78 \\ \cmidrule{2-6} 
         & X-Ray & Chest & CheXlocalize & Atelectasis, Cardiomegaly, Consolidation, Edema, Enlarged Cardiomediastinum, Lung Lesion, Lung Opacity, Pleural Effusion, Pneumothorax, and Support Devices & 1279/446/452 \\ \midrule
        \multirow{2}{*}{Open Domain} & Street Scenes & N/A & Cityscapes & Bicycle, Building, Bus, Car, Fence, Motorcycle, Person, Pole, Rider, Road, Sidewalk, Sky, Terrain, Traffic Light, Traffic Sign, Train, Truck, Vegetation, Wall & 34723/6005/- \\
         & Diverse Scenes & N/A & PascalVOC & Aeroplane, Bicycle, Bird, Boat, Bottle, Bus, Car, Cat, Chair, Cow, Dining table, Dog, Horse, Motorbike, Person, Potted plant, Sheep, Sofa, Train, TV Monitor & 2170/2148/- \\ \bottomrule 
    \end{tabular}%
    }
\end{table}

As open-domain datasets, we choose Cityscapes \cite{cordts2016cityscapes} and PASCAL VOC 2012 \cite{everingham2012pascal} datasets consisting of $19$ and $20$ classes, respectively.
The significance of using a natural-domain dataset is to evaluate the performance of various prompt-tuning methods in the same domain as the VLSMs were pretrained on.
Also, we can observe how the context learners perform when the number of classes ($\approx 20$) is higher compared to a maximum of $10$ in the above-mentioned medical datasets.
A detailed description of the datasets selected for this study is shown in \cref{tab:dataset_info}.

\begin{figure}[t]
    \centering
    \begin{subfigure}{0.5\textwidth}
        \includegraphics[width=\linewidth]{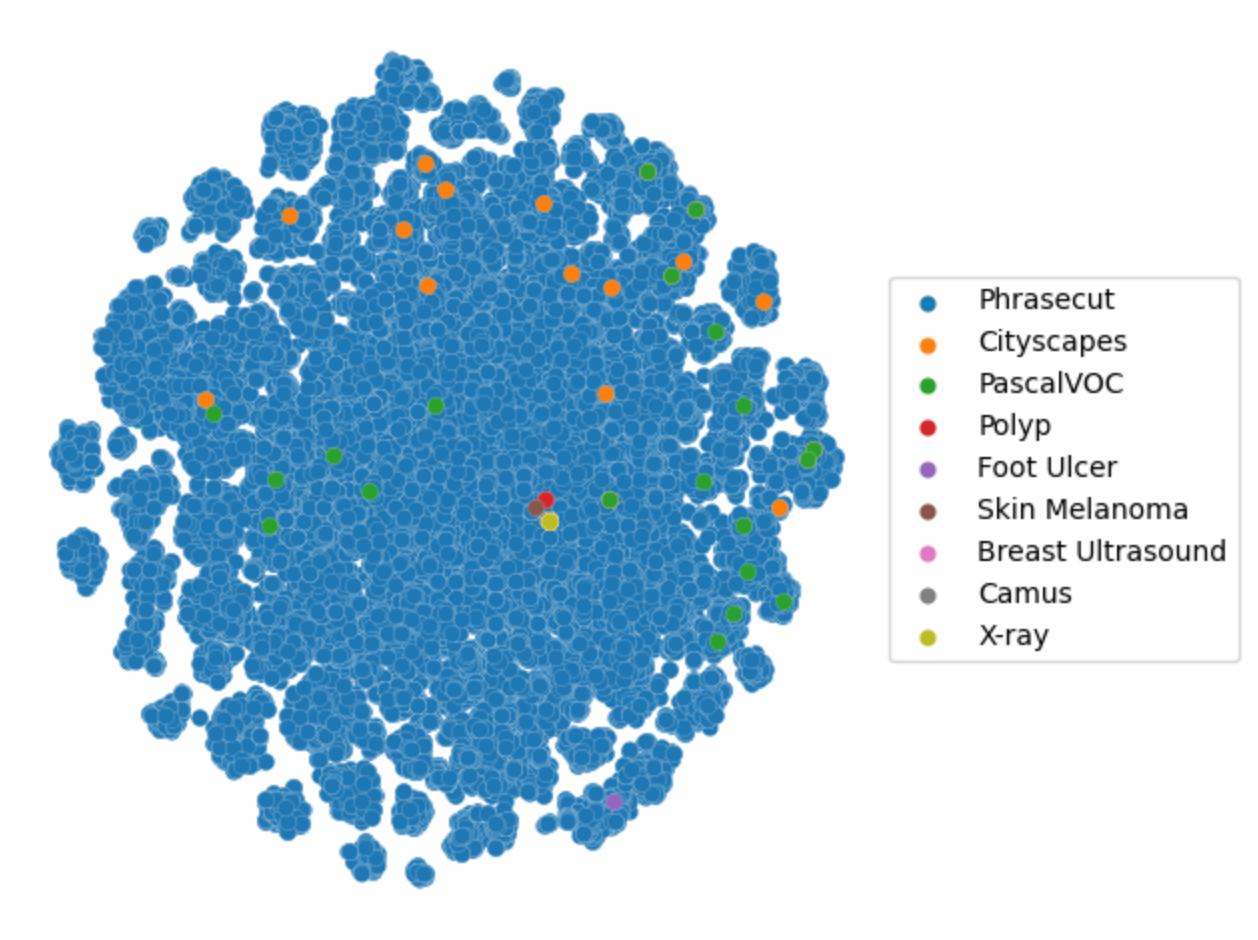}
        \caption{t-SNE of the Phrases}
        \label{fig:phrase_tsne}
    \end{subfigure}%
    \begin{subfigure}{0.5\textwidth}
        \includegraphics[width=\linewidth]{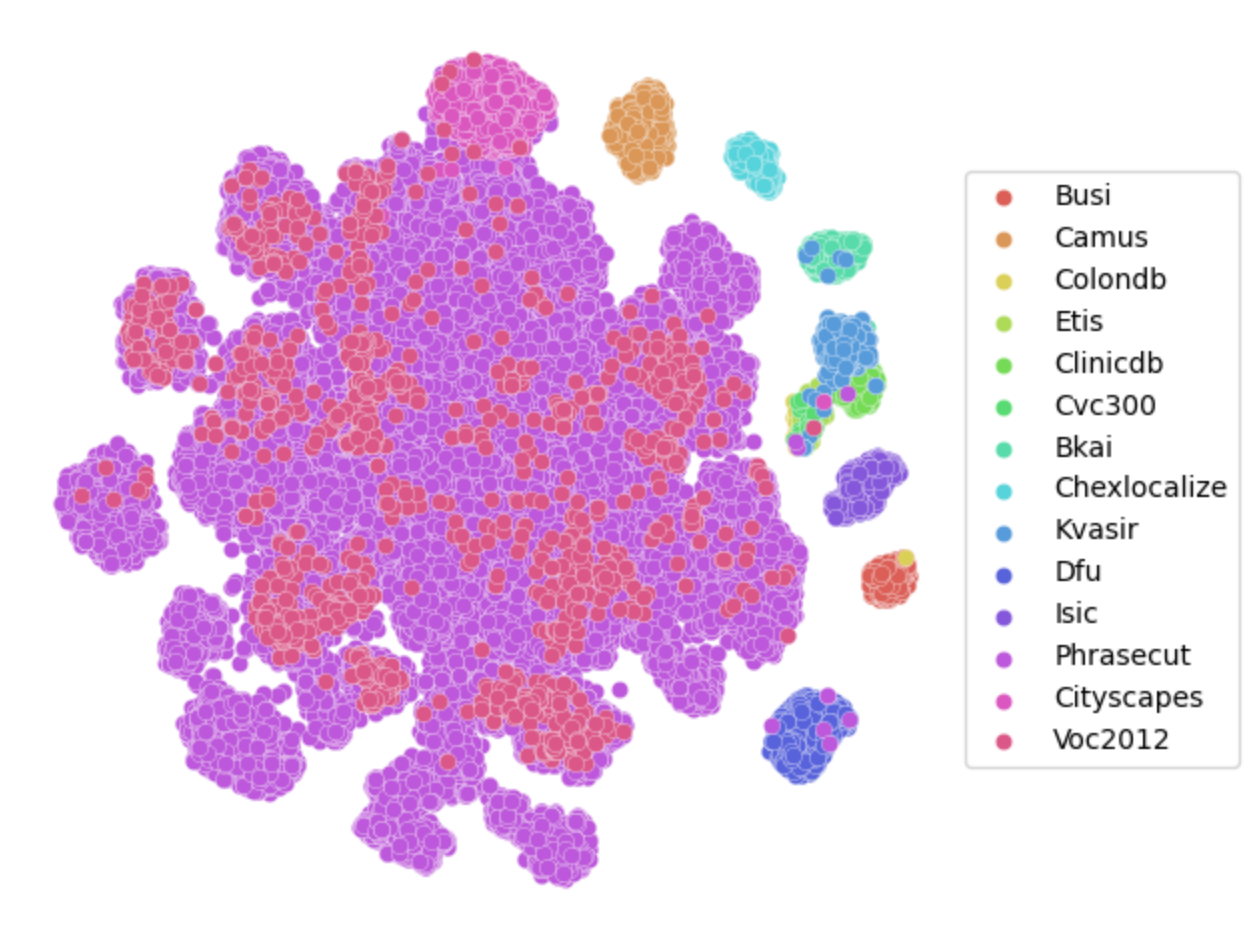}
        \caption{t-SNE of Images}
        \label{fig:image_tsne}
    \end{subfigure}
    \caption{
    t-SNE \cite{van2008visualizing} plots of phrases and images of all the datasets.
    Here, Phrasecut \cite{wu2020phrasecut} is the dataset on which CLIPSeg was pretrained, Cityscapes \cite{cordts2016cityscapes} and Pascal VOC2012 \cite{everingham2012pascal} are the open-domain datasets, while others correspond to the medical domain.
    In \cref{fig:image_tsne}, we can see the overlap of clusters for the open domain datasets, and the medical datasets have formed separate small clusters.
    }
    \label{fig:tsne}
\end{figure}

To analyze the domain shift between open-domain and medical datasets, we have plotted t-SNE \cite{van2008visualizing} of CLIP's text and vision embeddings for all datasets in \cref{fig:tsne}.
In addition to the datasets mentioned in \cref{sec:datasets}, we have also plotted the phrases in the Phrasecut dataset \cite{wu2020phrasecut}, the one used to pretrain CLIPSeg.
The figure has fewer text embeddings than image embeddings because many images correspond to a single prompt in the images in a dataset and the endoscopy datasets share the same foreground class (polyp).
Since the number of phrases for the semantic segmentation datasets is insignificant in comparison to the referring image segmentation dataset like PhraseCut, we cannot draw many conclusions from \cref{fig:phrase_tsne}.
However, in \cref{fig:image_tsne}, we can see the image embeddings for the medical datasets forming their clusters, separate from that of the open domain datasets.
Similarly, the image embeddings of the open domain datasets overlap and the datasets with polyp as their foreground class are in the same cluster (center right).
This illustrates a significant domain shift in CLIP's image embeddings from the dataset in which VLSMs were trained, to the medical datasets, thus, image embeddings require more adjustment to the domain shift.

\subsection{Implementation Details}
\label{sec:implementation_details}

Before feeding the images to the models, we scaled and normalized the images in the original resolution of the respective models viz. $416 \times 416$ for CRIS \cite{wang2022cris} and $352 \times 352$ for CLIPSeg \cite{luddecke2022image} using the bicubic interpolation.
Additionally, for the training split, we slightly augmented the images by randomly scaling the images in the range of $2\%$, translating $2\%$, and rotating $5^\circ$ using the Albumentations library \cite{buslaev2020albumentations}.
Additionally, the brightness and contrast of the images are augmented in the range of $10\%$.

We used 16-bit floating-point mixed-precision training in the NVIDIA GPUs using AdamW \cite{loshchilov2018decoupled} optimizer with an effective batch size of $32$ for all the experiments in this paper.
For the loss function, we used a combined loss of Dice Loss \cite{milletari2016v} and Binary Cross Entropy loss.
Mathematically, the loss function used can be shown as follows.
\[
    \mathcal{L} = \lambda_d \mathcal{L}_d + \lambda_{ce} \mathcal{L}_{ce}
\]
Here, $\lambda_d$ and $\lambda_{ce}$ were chosen to be $1$ and $0.2$, respectively for all the datasets.

\begin{table}[b]
    \centering
    \caption{The search space for the hyperparameter search.}
    \label{tab:hyperparameter_search}
    \begin{tabular}{llll}
        \toprule
        \textbf{Hyperparameter}         & \textbf{Search Space}         & \textbf{Applicable for} & \textbf{Space Type} \\ \midrule
        Learning rate                   & $[10^{-5}, 5 \times 10^{-3}]$ & ALL                     & Log                 \\
        Weight decay                    & $[10^{-5}, 0.01]$             & ALL                     & Log                 \\
        Prompt depth                    & $[1, 11]$                     & ALL                     & Integer             \\
        Intermediate dimension                & $\{32, 64, 96, 128\}$         & CoCoOp, Maple           & Choice              \\
        Use LORA                        & $\{true, false\}$             & CoCoOp, Maple           & Choice              \\
        Transformer: Number of Heads     & $\{16, 20, 32\}$              & Shared Attention        & Choice              \\
        Transformer: Dropout Probability & $[0.1, 0.55]$                 & Shared Attention        & Linear              \\
        Transformer: Feed-Forward Dim    & $\{1280, 1420\}$              & Shared Attention        & Choice              \\
        Transformer: LayerNorm First          & $\{true, false\}$             & Shared Attention        & Choice              \\
        Shared Space Dimension                & $\{32, 64\}$                  & Shared Separate         & Choice              \\ \bottomrule         
    \end{tabular}
\end{table}

As Jia \etal \cite{jia2022visual} shown that the same set of hyperparameters like learning rate and weight decay may not be optimal for all the datasets, we performed hyperparameter sweeps using TPESampler \cite{watanabe2023tree} from Optuna \cite{akiba2019optuna} library.
We ran each experiment $20$ times with the search space for each parameter shown in \cref{tab:hyperparameter_search}.
Although CoOp \cite{zhou2022learning} and CoCoOp \cite{zhou2022conditional} have proposed to use prompt depth of $1$, we wanted to observe how much can we benefit from increasing the prompt depth in those textual tuning techniques and to make the comparison across the prompting techniques fair and uniform, we have kept the prompt depth as a hyperparameter for all the context learners.

\section{Results and Discussion}

We report the results for each viable combination of models and context learns under the search space defined by \cref{tab:hyperparameter_search} corresponding to the best set of hyperparameters in \cref{tab:results}.
We explain the results shown in the table in subsequent subsections.


\begin{table}[b]
    \centering
    \caption{
    The dice scores for each viable combination of VLSMs, and context learners, and end-to-end(E2E) finetuning for all datasets.
    Each cell corresponds to the best dice score on each dataset after searching $20$ times using the search space defined by \cref{tab:hyperparameter_search}.  
    }
    \label{tab:results}
    \resizebox{\textwidth}{!}{%
    \begin{tabular}{@{}llccccccccccc@{}}
        \toprule
        \multicolumn{2}{r}{Datasets $\rightarrow$} & \multicolumn{5}{c}{Non-Radiology} & \multicolumn{3}{c}{Radiology} &  \multicolumn{2}{c}{Open Domain}& Overall \\ \cmidrule(r){1-2}\cmidrule(lr){3-7}\cmidrule(lr){8-10}\cmidrule(lr){11-12} \cmidrule(l){13-13}
        Model & Context Learner & BKAI & ClinicDB & Kvasir & DFU & ISIC & BUSI & CAMUS & Chexlocalize  & Cityscapes&PascalVOC& Mean \smallStd{Std}\\ \midrule \midrule
        CRIS & \multirow{2}{*}{E2E finetune} & 92.40 & 91.69 & 91.39 & 76.13 & 91.94 & 69.31 & 91.09 & 62.57 & 63.50 & 75.73 & \multirow{2}{*}{-}\\
        CLIPSeg &  & 86.47 & 88.74 & 89.51 & 73.24 & 92.12 & 64.32 & 88.85 & 59.56 & 59.43 & 78.88 & \\ \midrule
        \multirow{6}{*}{CLIPSeg} & Maple & $84.82$ & $\mathbf{90.61}$ & $87.25$& $\mathbf{72.68}$ & $\mathbf{92.10}$& $80.99$& $88.95$& $58.04$  & $\mathbf{56.70}$ & $\mathbf{79.48}$ & $79.16 \smallStd{12.2}$\\
        & Shared Sep & $86.06$ & $90.30$ & $88.02$ & $71.71$ & $92.03$ & $81.71$& $89.29$ & $\mathbf{58.95}$  & $\mathbf{56.70}$ & $79.46$ & $79.42 \smallStd{12.22}$\\
        & Shared Attn & $84.63$ & $88.43$ & $88.07$ & $68.56$ & $91.77$ & $78.22$& $83.08$ & $53.24$  & $55.58$ & $78.80$ & $77.04 \smallStd{12.91}$\\
        & VPT & $\mathbf{87.96}$ & $90.31$ & $\mathbf{89.03}$ & $71.76$ & $91.99$ & $\mathbf{82.45}$& $\mathbf{89.36}$ & $57.67$  & $56.51$ & $79.29$& $\mathbf{79.63 \smallStd{12.67}}$\\
        & CoCoOp & $56.78$ & $74.21$ & $75.63$ & $58.64$ & $89.56$ & $70.19$& $62.16$ & $42.77$  & $47.65$ & $75.66 $& $65.33 \smallStd{13.61}$\\
        & CoOp & $53.85$ & $66.59$ & $73.53$ & $54.64$ & $89.00$ & $67.28$& $58.92$ & $41.82$  & $47.92$ & $75.50$ & $62.91 \smallStd{13.49}$\\ \midrule
        \multirow{2}{*}{CRIS} & CoCoOp & $76.85$ & $85.09$ & $82.12$ & $58.24$ & $86.87$ & $72.95$& $81.96$ & $51.28$  & $44.88$ & $72.46$ & $71.27 \smallStd{14.03}$\\
        & CoOp & $75.55$ & $76.73$ & $80.38$ & $57.18$ & $86.10$ & $75.46$& $79.23$ & $51.79$  & $46.50$ & $71.72$ & $70.06 \smallStd{12.69}$\\ \midrule
        \multicolumn{12}{r}{Average across Methods} & $73.10 \smallStd{6.24}$\\ \bottomrule
    \end{tabular}%
    }
\end{table}

\subsection{Choice of Prompt Tuning Technique}

From \cref{tab:results}, we can see that the textual prompt tuning methods viz. CoCoOp and CoOp perform significantly worse than other prompt tuning techniques.
The table shows that although Maple \cite{khattak2023maple} has the highest dice scores for some datasets, VPT \cite{jia2022visual} has the highest average dice score across the test datasets despite not tuning the textual prompts.
We hypothesize that since we used the same number of search trials for all the context learners, finding the optimal set of hyperparameters for VPT is easier as it has fewer hyperparameters than the multimodal variants (see \cref{tab:hyperparameter_search}).
Also, this could have occurred because, as shown in \cref{fig:tsne}, there is a significant domain shift from the open domain to the medical domain in the image domain only.
This premise is supported by the observation that VPT performs better than the multimodal variants in the medical domain but the multimodal variants perform better than VPT in the open domain datasets (see \cref{tab:hyperparameter_search}).
From this observation, we can infer that VPT is a good starting point when you have a dataset different from the ones on which the VLSMs were pretrained.

\subsection{Importance of Prompt Initialization}

As shown by Zhou \etal \cite{zhou2022learning,zhou2022conditional}, a good initialization of the prompts leads to a better performance.
Rather than randomly initializing the first prompt to CLIP's text encoder, they found initializing it with the embeddings of ``a photo of a'' led to a better performance.
So, for the context learners where the prompts are initialized in the textual space viz. CoOp, CoCoOp, and Maple, the first prompt is initialized as mentioned above.
We hypothesize that the ability to initialize the context vectors in the text space of the CLIP's text encoder in Maple \cite{khattak2023maple} could be the reason why Maple performed better than the other multimodal prompt tuning techniques in some datasets.
We trained Maple on CLIPSeg by initializing the context vectors from a Gaussian distribution ($\mu = 0, \sigma = 0.02$).
From \cref{tab:maple_random_init}, we can see that the performance of Maple decreases for most of the dataset when the context vectors for the first depth are initialized at random, which supports our hypothesis.
Additionally, the performance of different prompt tuning methods, is comparable for most of the datasets with end-to-end fine-tuning (Table \ref{tab:results}, scores for the medical datasets are extracted from Poudel \etal \cite{poudel2023exploring}  and the rest are trained with the same set of hyperparameters.).

\begin{table}[t]
    \centering
    \caption{
    Performance of Maple when the first vector is initialized from Gaussian distribution vs `a photo of a'.
    }
    \label{tab:maple_random_init}
    \resizebox{\textwidth}{!}{%
    \begin{tabular}{@{}lccccccccccc@{}}
        \toprule
         \multicolumn{1}{r}{Datasets $\rightarrow$} &  \multicolumn{5}{c}{Non-Radiology}&  \multicolumn{3}{c}{Radiology}& \multicolumn{2}{c}{Open Domain}& Overall\\
        \cmidrule(r){1-1}\cmidrule(lr){2-7}\cmidrule(lr){8-9}\cmidrule(lr){10-11}\cmidrule(l){12-12}
         Init $\downarrow$ &  BKAI&  ClinicDB&  Kvasir&  DFU&  ISIC&  BUSI&  CAMUS&  Chexlocalize& Cityscapes& PascalVOC& Mean {$\pm$ \scriptsize Std}\\ \midrule
         Gaussian & $83.96$ & $88.34$ & $\mathbf{88.65}$ & $70.14$ & $92.05$ & $\mathbf{81.25}$ & $88.25$ & $57.27$ & $56.62$ & $\mathbf{79.50}$ & $78.6 \smallStd{12.31}$\\
         a photo of a & $\mathbf{84.82}$ & $\mathbf{90.61}$ & $87.25$ & $\mathbf{72.68}$ & $\mathbf{92.10}$ & $80.99$ & $\mathbf{88.95}$ & $\mathbf{58.04}$ & $\mathbf{56.70}$ & $79.48$ & $\mathbf{79.16 \smallStd{12.2}}$\\
         \bottomrule
    \end{tabular}%
    }
\end{table}

\subsection{Importance of Tuning Last Layer for Segmentation}

\begin{table}[b]
    \centering
    \caption{
    The dice scores for each viable combination of VLSMs and context learners for all the datasets when a new learnable last upsampling layer is not introduced.
    }
    \label{tab:no_last_layer}
    \resizebox{\textwidth}{!}{%
    \begin{tabular}{@{}llccccccccccc@{}}
        \toprule
        \multicolumn{2}{r}{Datasets $\rightarrow$} & \multicolumn{5}{c}{Non-Radiology} & \multicolumn{3}{c}{Radiology} &  \multicolumn{2}{c}{Open Domain}& Overall \\ \cmidrule(r){1-2}\cmidrule(lr){3-7}\cmidrule(lr){8-10}\cmidrule(lr){11-12} \cmidrule(l){13-13}
        Models & Context Learner & BKAI & ClinicDB & Kvasir & DFU & ISIC & BUSI & CAMUS & Chexlocalize & Cityscapes&PascalVOC& Mean {$\pm$ \scriptsize Std} \\ \midrule
        \multirow{6}{*}{CLIPSeg} & Maple & $81.56$ & $85.55$ & $86.45$ & $66.38$ & $91.20$ & $80.95$ & $84.96$ & $55.96$ & $53.16$ & $76.95$ & $76.31 \smallStd{12.58}$\\
        & Shared Sep & $80.45$ & $83.31$ & $86.07$ & $65.46$ & $91.00$ & $78.81$ & $85.71$ & $57.72$ & $53.06$ & $76.72$ & $75.83 \smallStd{12.14}$\\
        & Shared Attn & $80.21$ & $85.89$ & $85.62$ & $62.23$ & $90.91$ & $76.27$ & $84.30$ & $55.77$ & $51.86$ & $76.50$ & $74.96 \smallStd{12.92}$\\
        & VPT & $81.57$ & $85.22$ & $86.64$ & $65.06$ & $91.33$ & $78.93$ & $85.32$ & $57.22$ & $52.65$ & $76.18$ & $76.01 \smallStd{12.56}$\\
        & CoCoOp & $55.75$ & $73.03$ & $75.54$ & $57.90$ & $89.55$ & $70.60$ & $60.23$ & $42.61$ & $47.91$ & $75.85$ & $64.9 \smallStd{13.71}$\\
        & CoOp & $53.80$ & $66.56$ & $73.29$ & $54.53$ & $88.98$ & $67.85$ & $57.93$ & $41.06$ & $47.91$ & $75.47$ & $62.74 \smallStd{13.65}$\\ \midrule
        \multirow{2}{*}{CRIS} & CoCoOp & $72.63$ & $77.19$ & $81.69$ & $53.05$ & $84.51$ & $62.62$ & $75.71$ & $47.45$ & $43.20$ & $70.04$ & $66.81 \smallStd{13.8}$\\
        & CoOp & $72.14$ & $75.73$ & $79.44$ & $52.02$ & $83.44$ & $64.53$ & $75.03$ & $48.71$ & $44.03$ & $70.03$ & $66.51 \smallStd{13.01}$\\ \midrule
        \multicolumn{12}{r}{Average across Methods} & $70.51 \smallStd{5.4}$\\ \bottomrule
    \end{tabular}%
    }
\end{table}

As mentioned in \cref{sec:method}, we have used a learnable upsampler as a residual connection to the decoder's last layer.
To study the importance of the layer, we conduct an ablation by removing the learnable layer, keeping encoders and decoder frozen; the results from the ablation are shown in \cref{tab:no_last_layer}.
From the table, we can see that the dice score has decreased for most of the cases, reducing the overall average dice score by $2.59$.
This shows the advantage of the learnable upsampler for segmentation.

\subsection{Importance of Prompt Depth}

\begin{figure}[t]
    \centering
    \begin{subfigure}{0.5\textwidth}
        \includegraphics[width=\linewidth]{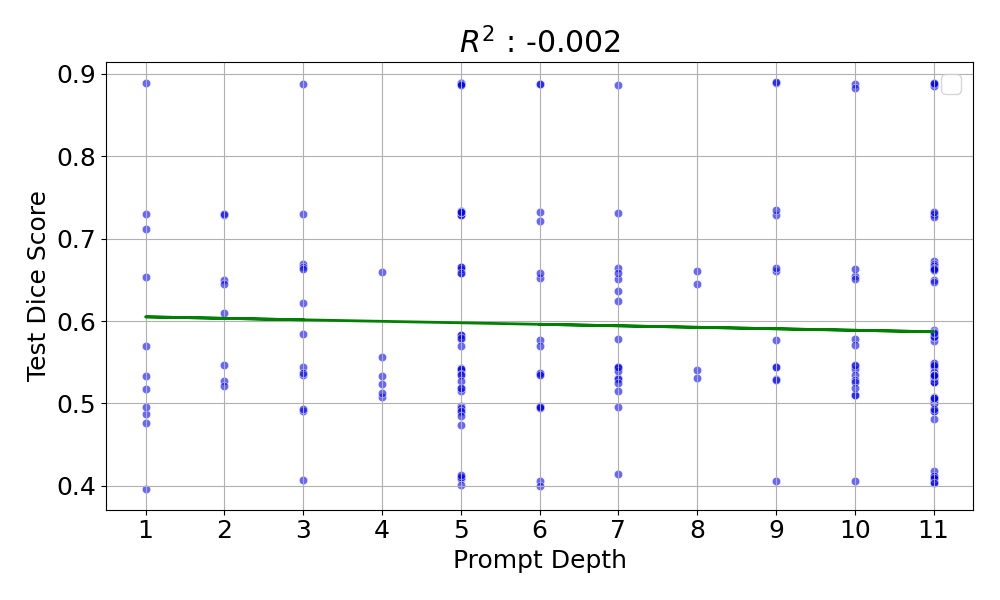}
        \caption{CoOp CLIPSeg}
        \label{fig:prompt_depth_coop_clipseg}
    \end{subfigure}%
    \begin{subfigure}{0.5\textwidth}
        \includegraphics[width=\linewidth]{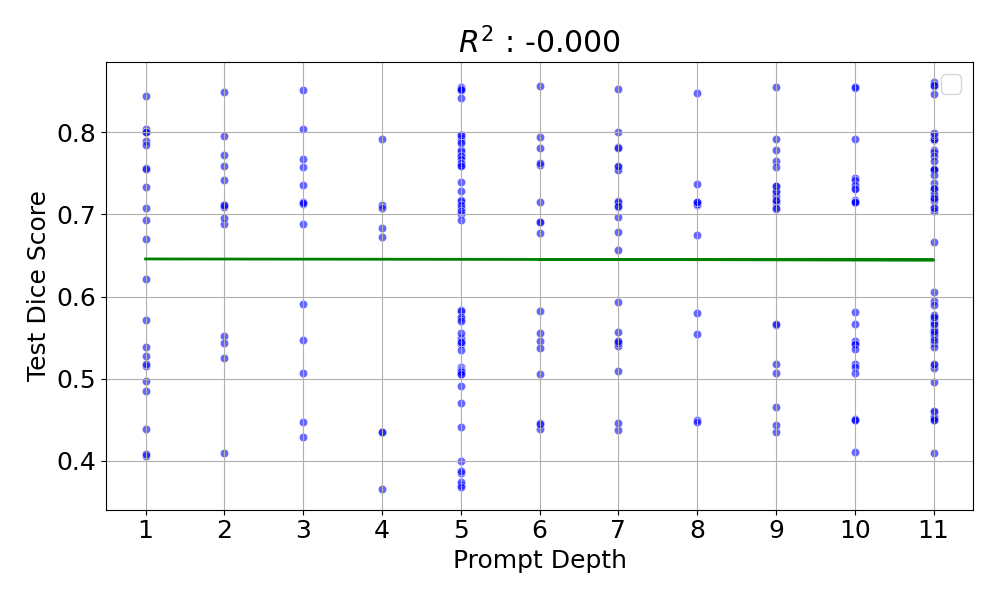}
        \caption{CoOp CRIS}
        \label{fig:prompt_depth_coop_cris}
    \end{subfigure}
    \begin{subfigure}{0.5\textwidth}
        \includegraphics[width=\linewidth]{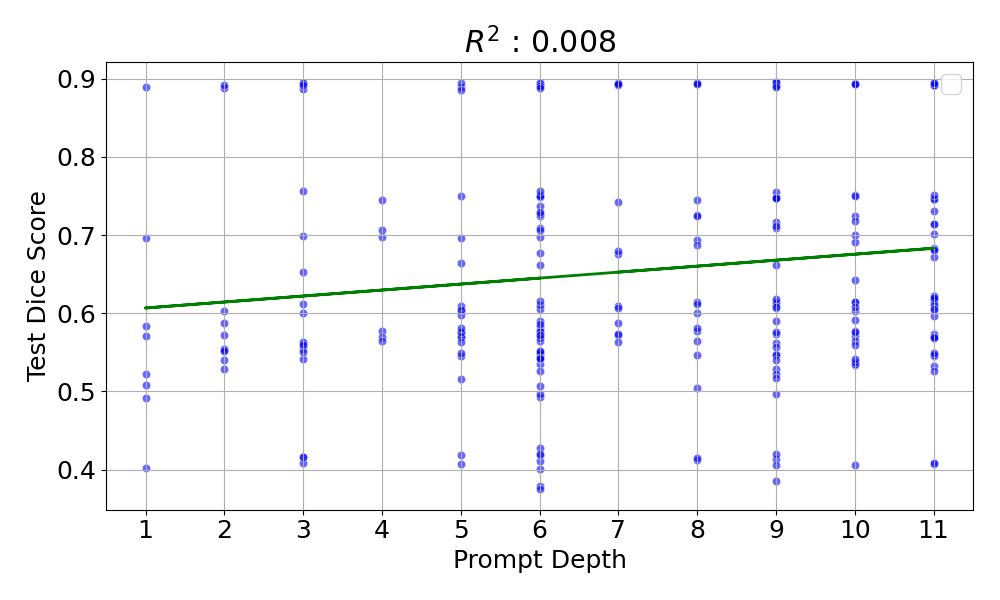}
        \caption{CoCoOp CLIPSeg}
        \label{fig:prompt_depth_cocoop_clipseg}
    \end{subfigure}%
    \begin{subfigure}{0.5\textwidth}
        \includegraphics[width=\linewidth]{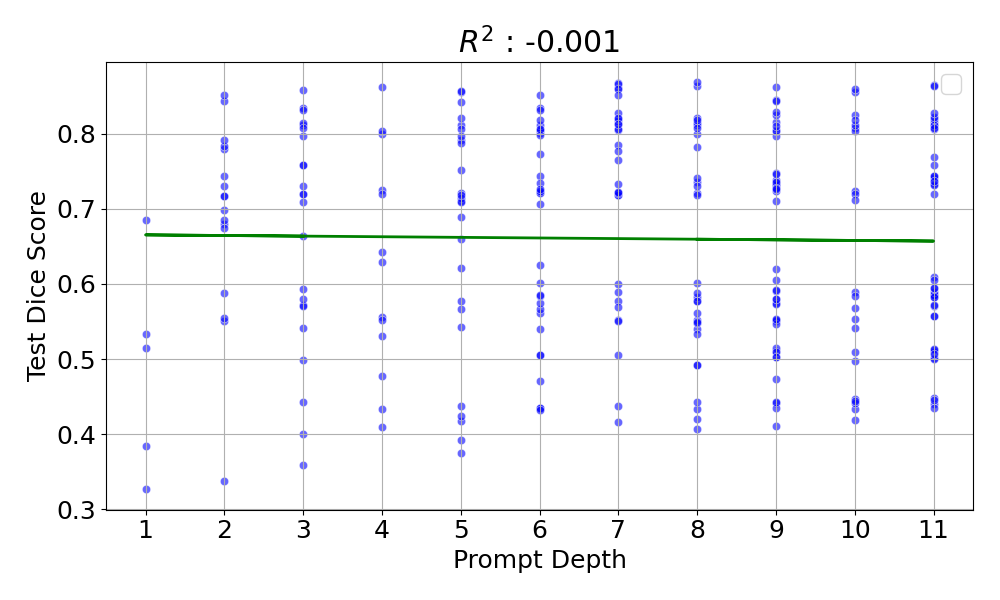}
        \caption{CoCoOp CRIS}
        \label{fig:prompt_depth_cocoop_cris}
    \end{subfigure}
    \caption{Test Dice \vs Prompt Depth for Textual Tuning of all Datasets}
    \label{fig:prompt_depth_imp}
\end{figure}

Since prompt depth has shown to be an important hyperparameter of prompt tuning \cite{jia2022visual,khattak2023maple,zang2022unified}, we ran the experiments for various prompt depths as mentioned in \cref{sec:implementation_details}.
We have plotted the test dice score versus the prompt depth for various context learners for all the datasets in \cref{fig:prompt_depth_imp}.
The reason to include the dice score for all the datasets in the same graph is to observe if each prompt learning technique improves in general on increasing the prompt depth.
From \cref{fig:prompt_depth_coop_clipseg,fig:prompt_depth_coop_cris,fig:prompt_depth_cocoop_clipseg,fig:prompt_depth_cocoop_cris}, we can see that the textual context learners do not benefit significantly on increasing the prompt depth.

\subsection{Learning Rate and Weight Decay}

Similar to Jia \etal \cite{jia2022visual}, we also observed that no single learning rate and weight decay pair performs the best for all the dataset and context learners (see Appendix).
So, we need to perform some sort of hyperparameter sweep to obtain the best set of configurations for the downstream dataset of choice.

\section{Limitations}
This work focuses on using VLSMs pretrained on segmentation tasks and evaluates their performance on downstream tasks.
Since the classes used to train the models may not be the same as the ones used to evaluate the models in the downstream tasks, we chose VLSMs that output a binary mask, and the class to segment is provided through the text encoder.
We refer to the VLSMs whose parameters do not need to be changed as the classes in downstream tasks differ from the pretrained task as \textit{class-agnostic VLSMs}.
Also, since the multimodal prompt tuning assumes the vision and text encoders run in parallel, one depending upon another, we restricted our study to VLSMs that use both encoders in (almost) parallel settings.
Due to these constraints, we bounded our scope of study to CLIPSeg \cite{luddecke2022image} and CRIS \cite{wang2022cris}, VLSMs which output binary segmentation masks and use CLIP's encoders for the segmentation tasks.

\section{Conclusion}

In this paper, we propose an open-source benchmark framework, \textit{TuneVLSeg}, to evaluate various unimodal and multimodal prompt tuning strategies on VLSMs for the segmentation task.
Although VLSMs other than the ones used in this paper are also available, we used only the class-agnostic VLSMs because of the difference in the classes between the pretrained and the downstream datasets.
Nevertheless, this can be extended to more VLSMs, if retaining the pretrained weights is not an issue, and more prompt tuning methodologies with minimum effort.
Besides, to present the effectiveness of our framework, we report the performance of $6$ prompt tuning methods on adapting $2$ CLIP-based VLSMs to $2$ natural domain and $8$ diverse medical domain segmentation datasets.
This work presents prompt tuning as an effective strategy to adapt pretrained foundational VLSMs for domain-specific segmentation tasks, with potential applications in clinical settings for medical image segmentation.

\subsubsection{\ackname} We would like to thank SUNY Korea for providing us with the computational resources required for this research project.

%
%
\bibliographystyle{splncs04}
\bibliography{references}

\appendix
\include{supplementary_core}

\end{document}

%% file: supplementary_core.tex
\section{How does prompt depth affect Dice Score?}

\begin{figure}[h]
    \centering
    \begin{subfigure}{0.5\textwidth}
        \includegraphics[width=\linewidth]{prompt_depth_coop_clipseg}
        \caption{CoOp CLIPSeg}
        \label{fig:prompt_depth_coop_clipseg}
    \end{subfigure}%
    \begin{subfigure}{0.5\textwidth}
        \includegraphics[width=\linewidth]{prompt_depth_coop_cris}
        \caption{CoOp CRIS}
        \label{fig:prompt_depth_coop_cris}
    \end{subfigure}
    \begin{subfigure}{0.5\textwidth}
        \includegraphics[width=\linewidth]{prompt_depth_cocoop_clipseg}
        \caption{CoCoOp CLIPSeg}
        \label{fig:prompt_depth_cocoop_clipseg}
    \end{subfigure}%
    \begin{subfigure}{0.5\textwidth}
        \includegraphics[width=\linewidth]{prompt_depth_cocoop_cris}
        \caption{CoCoOp CRIS}
        \label{fig:prompt_depth_cocoop_cris}
    \end{subfigure}
    \begin{subfigure}{0.5\textwidth}
        \includegraphics[width=\linewidth]{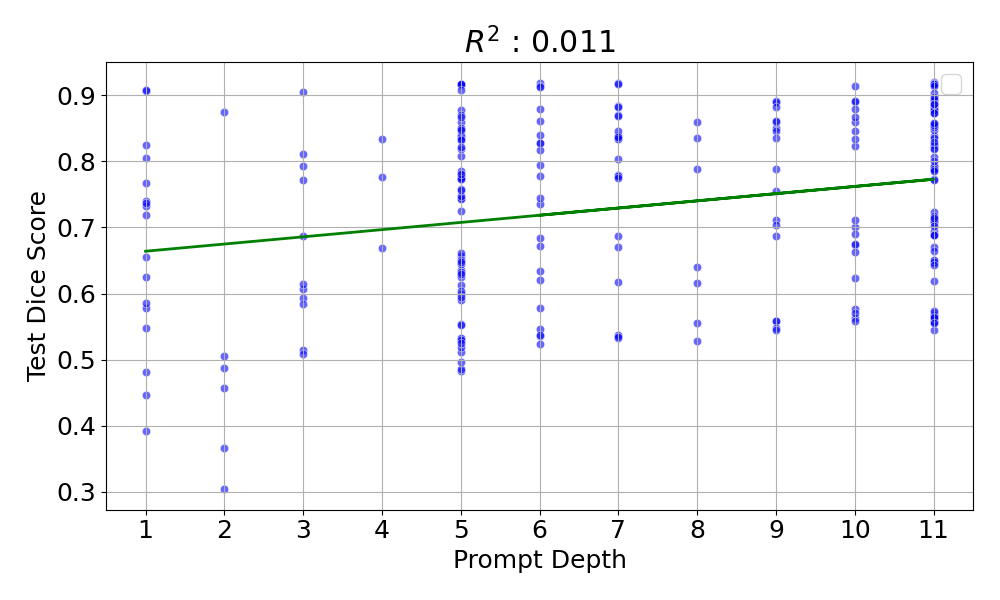}
        \caption{VPT CLIPSeg}
        \label{fig:prompt_depth_vpt_clipseg}
    \end{subfigure}%
    \begin{subfigure}{0.5\textwidth}
        \includegraphics[width=\linewidth]{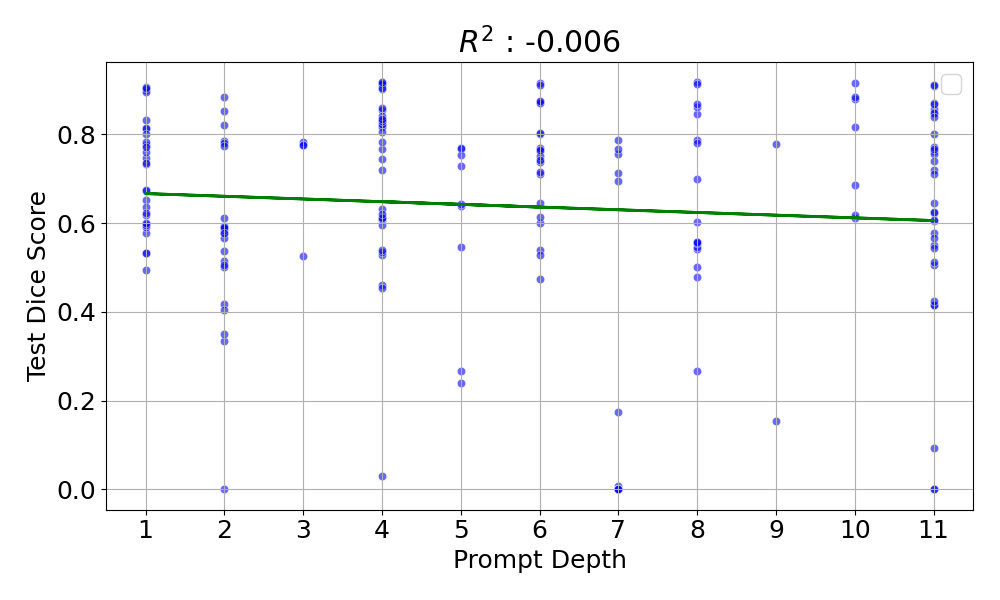}
        \caption{Shared Attention CLIPSeg}
        \label{fig:prompt_depth_shared_attn_clipseg}
    \end{subfigure}
    \begin{subfigure}{0.5\textwidth}
        \includegraphics[width=\linewidth]{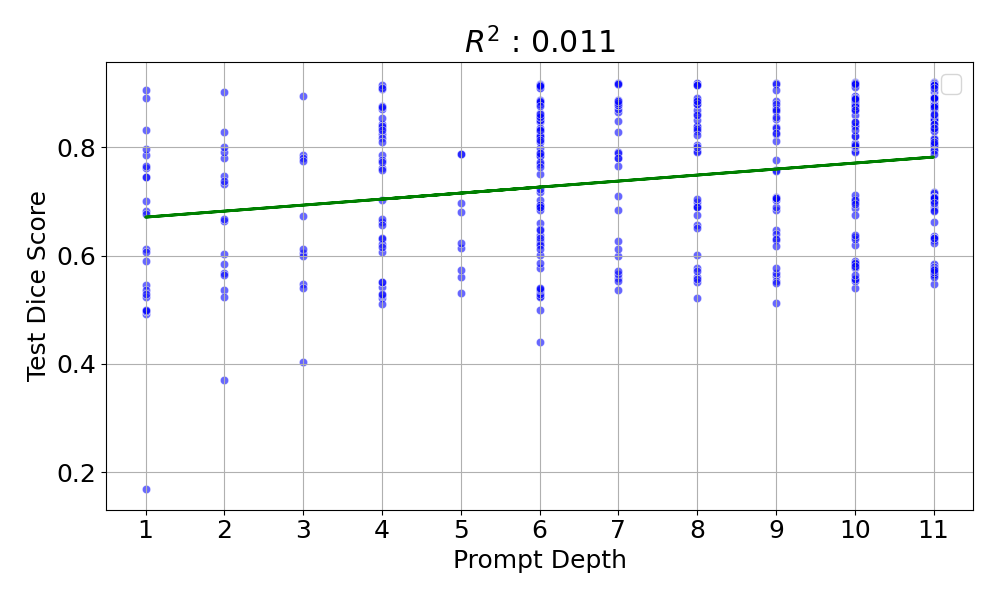}
        \caption{Shared Separate CLIPSeg}
        \label{fig:prompt_depth_shared_separate_clipseg}
    \end{subfigure}%
    \begin{subfigure}{0.5\textwidth}
        \includegraphics[width=\linewidth]{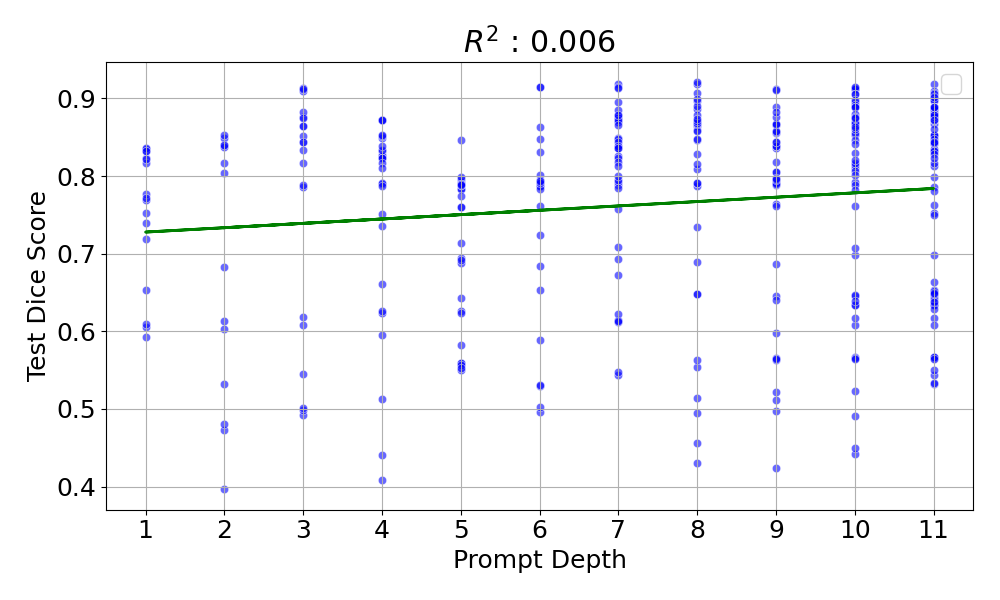}
        \caption{Maple CLIPSeg}
        \label{fig:prompt_depth_maple_clipseg}
    \end{subfigure}
    \caption{Test Dice \vs Prompt Depth for Textual Tuning for all Datasets}
    \label{fig:prompt_depth_imp}
\end{figure}

The scatter plot of test dice vs the prompt depth is shown in \cref{fig:prompt_depth_imp}.
We have also fitted a linear regression for the data and have shown $R^2$ score for each context learner.
There isn't a strong correlation between test dice and the prompt depth when the dice score, indicating that increasing prompt depth may not always increase the dice score for all the datasets and methods.

\section{How does learning rate and weight decay affect Dice?}

\begin{figure}[h]
    \centering
    \begin{subfigure}{0.5\textwidth}
        \includegraphics[width=\linewidth]{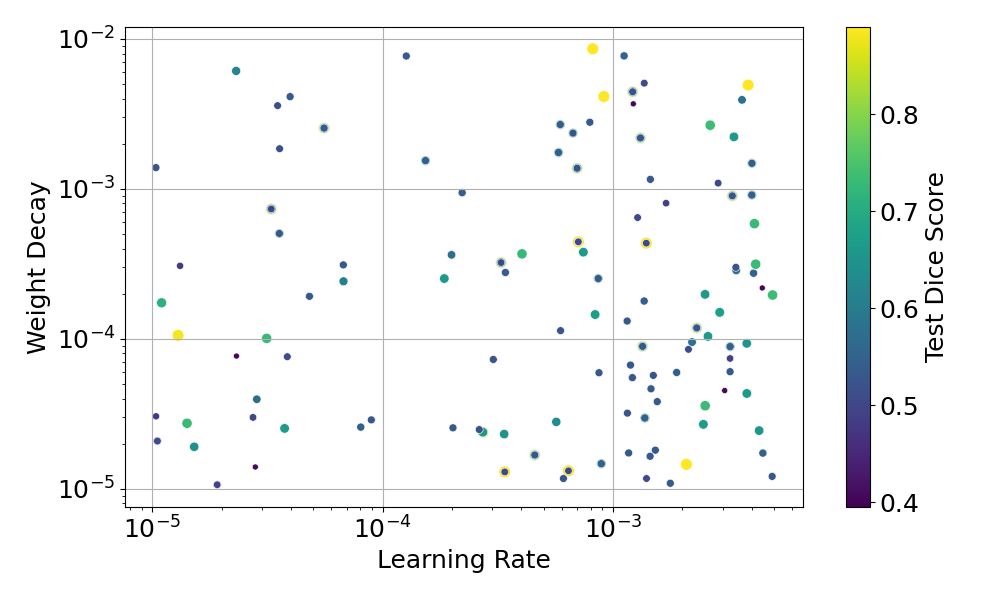}
        \caption{CoOp CLIPSeg}
        \label{fig:lr_wd_coop_clipseg}
    \end{subfigure}%
    \begin{subfigure}{0.5\textwidth}
        \includegraphics[width=\linewidth]{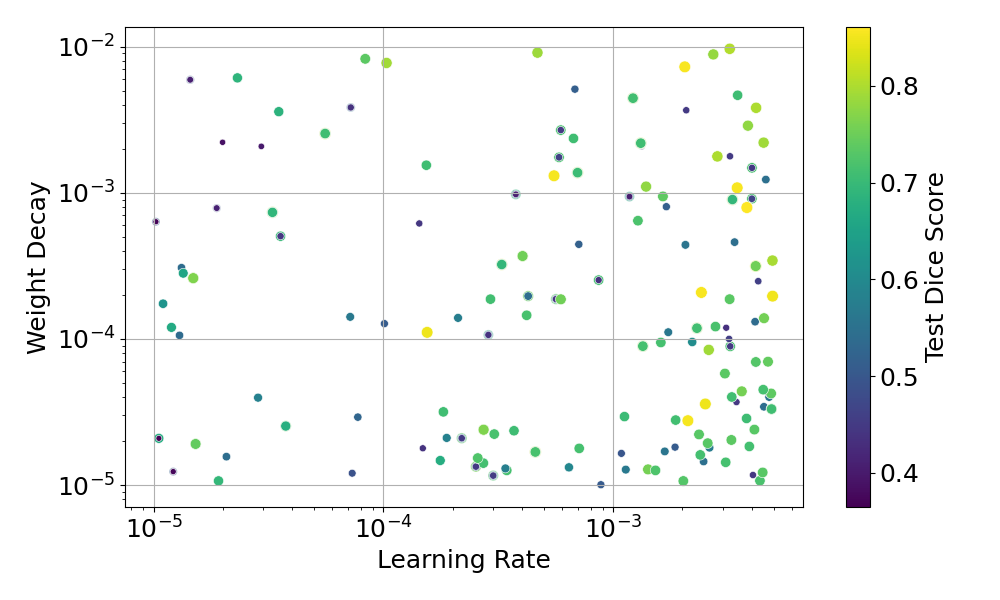}
        \caption{CoOp CRIS}
        \label{fig:lr_wd_coop_cris}
    \end{subfigure}
    \begin{subfigure}{0.5\textwidth}
        \includegraphics[width=\linewidth]{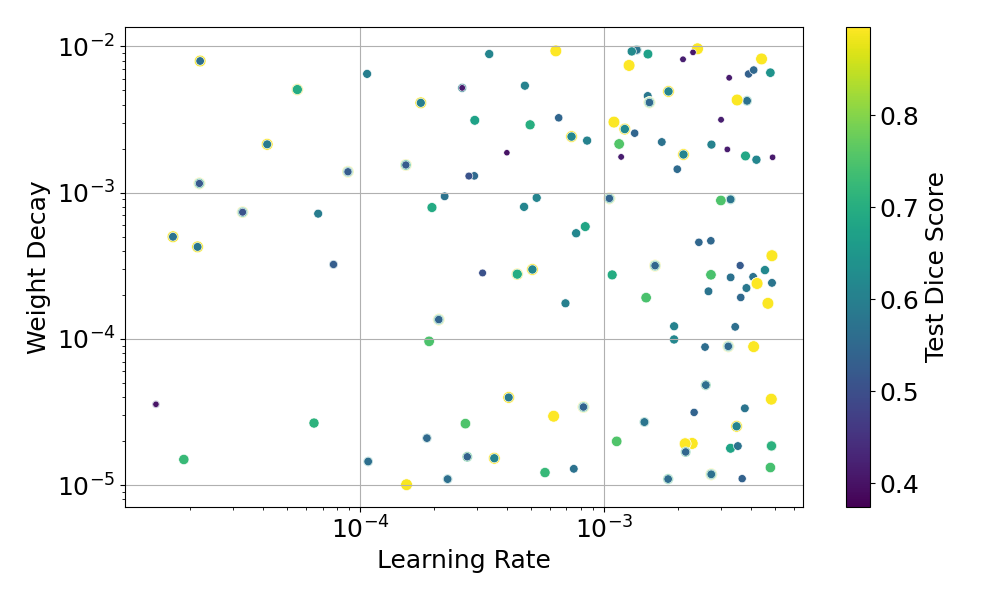}
        \caption{CoCoOp CLIPSeg}
        \label{fig:lr_wd_cocoop_clipseg}
    \end{subfigure}%
    \begin{subfigure}{0.5\textwidth}
        \includegraphics[width=\linewidth]{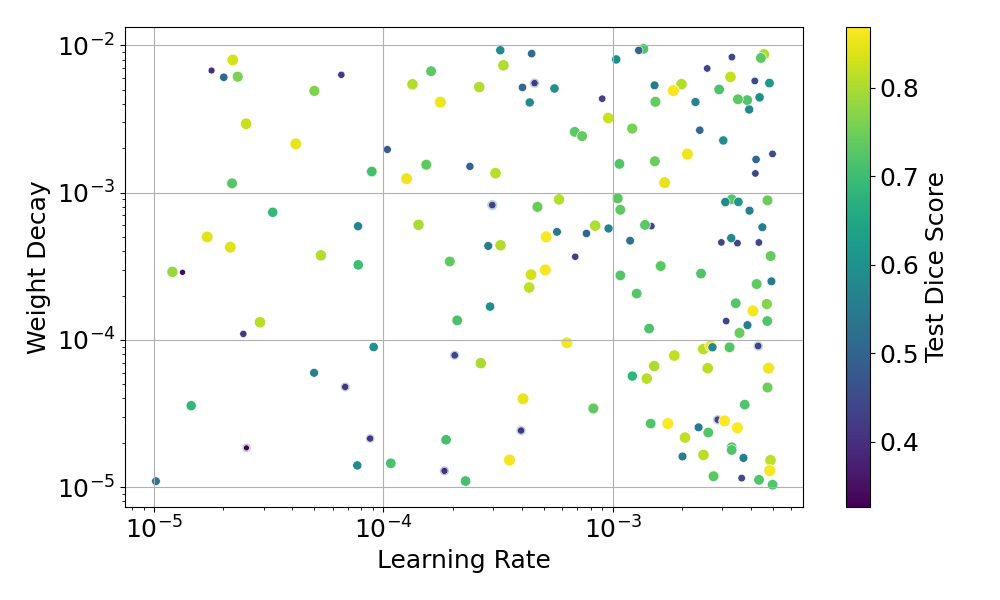}
        \caption{CoCoOp CRIS}
        \label{fig:lr_wd_cocoop_cris}
    \end{subfigure}
    \begin{subfigure}{0.5\textwidth}
        \includegraphics[width=\linewidth]{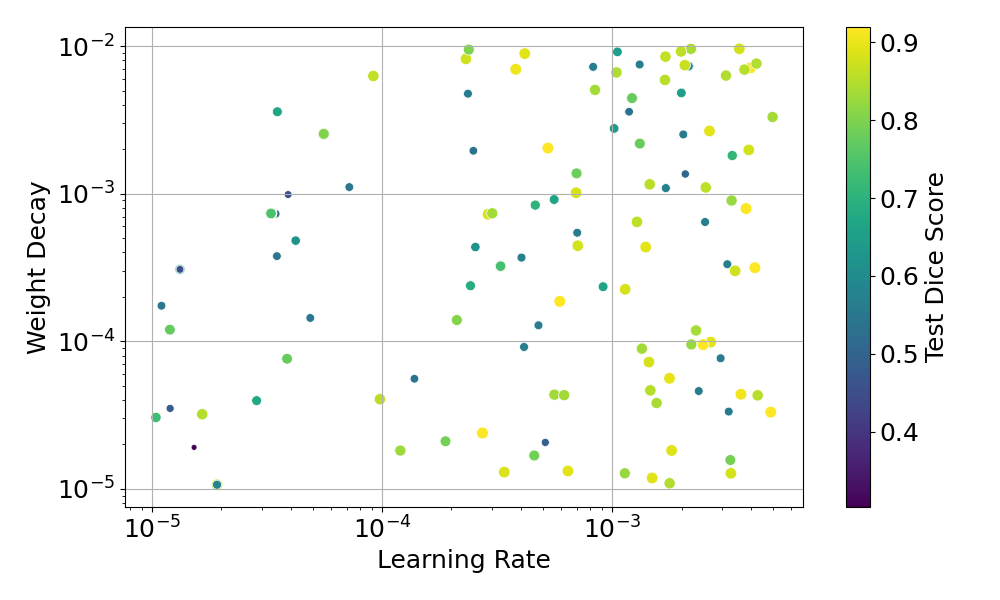}
        \caption{VPT CLIPSeg}
        \label{fig:lr_wd_vpt_clipseg}
    \end{subfigure}%
    \begin{subfigure}{0.5\textwidth}
        \includegraphics[width=\linewidth]{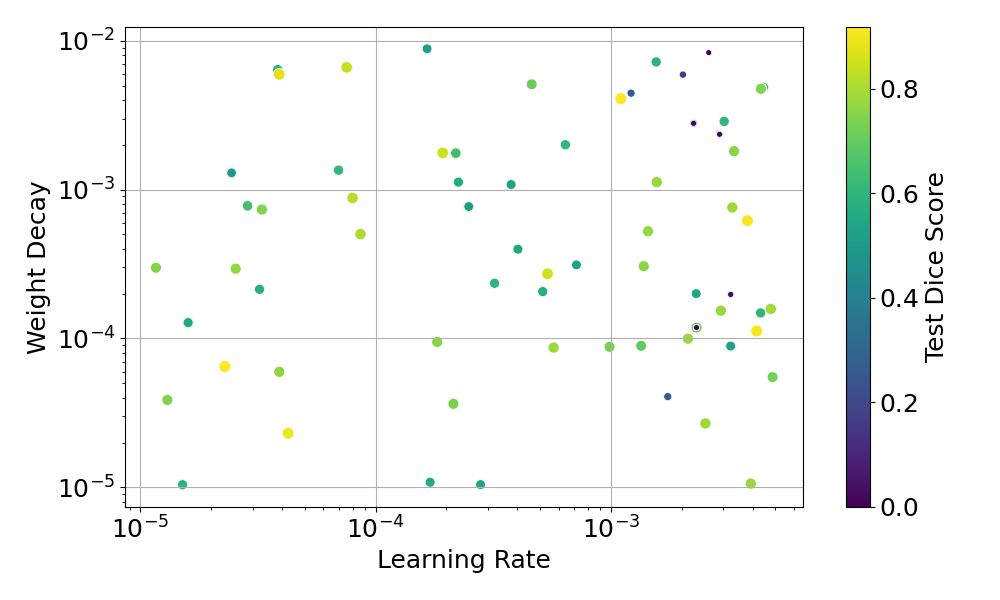}
        \caption{Shared Attention CLIPSeg}
        \label{fig:lr_wd_shared_attn_clipseg}
    \end{subfigure}
    \begin{subfigure}{0.5\textwidth}
        \includegraphics[width=\linewidth]{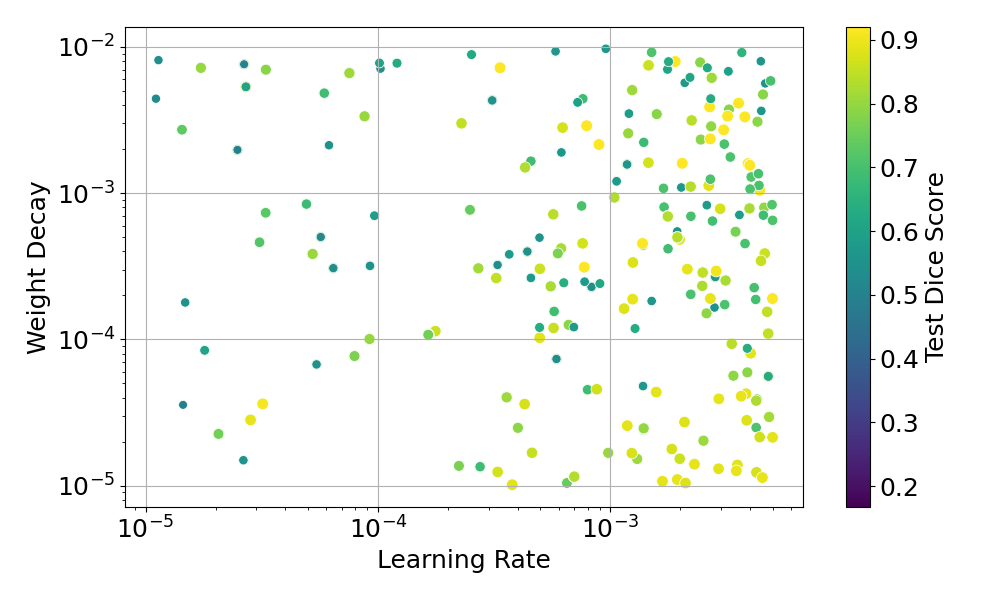}
        \caption{Shared Separate CLIPSeg}
        \label{fig:lr_wd_shared_separate_clipseg}
    \end{subfigure}%
    \begin{subfigure}{0.5\textwidth}
        \includegraphics[width=\linewidth]{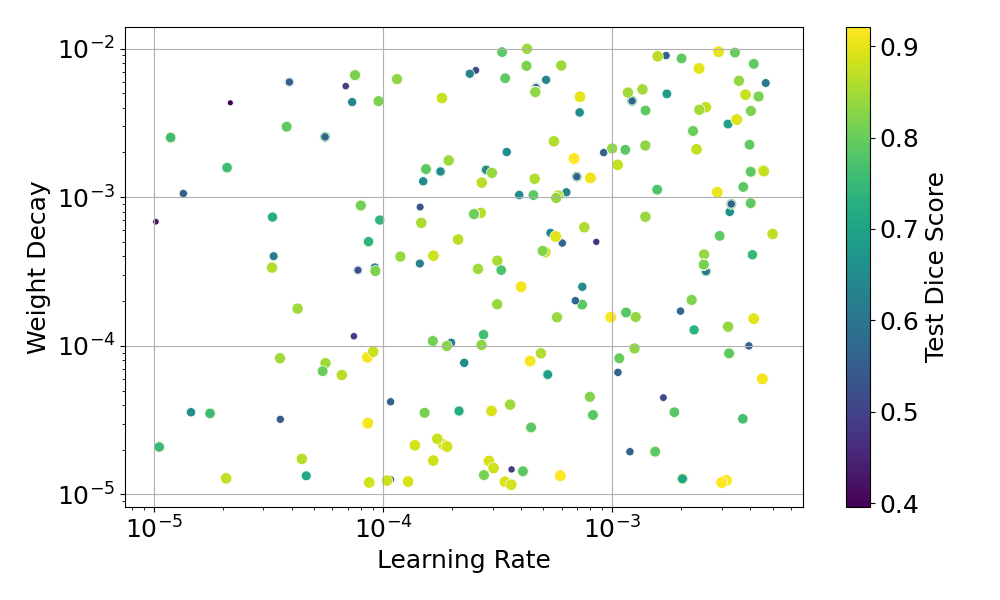}
        \caption{Maple CLIPSeg}
        \label{fig:lr_wd_maple_clipseg}
    \end{subfigure}
    \caption{Test Dice \vs Learning Rate \vs Weight Decay for all Datasets}
    \label{fig:lr_vs_weight_decay}
\end{figure}

The scatter plot showing the test dice on different weight decays and learning rates is shown in \cref{fig:lr_vs_weight_decay}.
Although there doesn't seem to be a direct between weight decay and test dice nor learning rate and test dice, a learning rate of around $10^{-3}$ seems to be a good starting point to train these context learners. 

